\documentclass[letterpaper, 10 pt, journal, twoside]{IEEEtran}

\usepackage{graphics} 
\usepackage{epsfig} 
\usepackage{bm}
\usepackage{xcolor}
\usepackage{amsmath} 
\usepackage{amssymb}  
\usepackage{cite}
\usepackage{caption} 
\usepackage{booktabs}
\usepackage{algorithm}
\usepackage{algorithmicx}
\usepackage{algpseudocode}
\usepackage{multirow}
\usepackage{hyperref}
\usepackage{relsize}

\title{Kernel-based Metrics Learning for Uncertain Opponent Vehicle Trajectory Prediction in Autonomous Racing}

\author{Hojin Lee, Youngim Nam, Sanghun Lee, and Cheolhyeon Kwon
\thanks{Manuscript received: May 30, 2024; Revised: September 1, 2024; Accepted: October 16, 2024.}
\thanks{This paper was recommended for publication by Editor Jens Kober upon evaluation of the Associate Editor and Reviewers' comments. This work was supported by the National Research Foundation of Korea(NRF) grants funded by the Korea government (MSIT) (RS-2024-00342930 and 2020R1A5A8018822). \textit{(corresponding author: Cheolhyeon Kwon)}}
\thanks{The authors are with the Department of Mechanical Engineering, Ulsan National Institute of Science and Technology, Ulsan, 44919, Republic of Korea 
        {\tt\footnotesize hojinlee@unist.ac.kr, nyi0944@unist.ac.kr, sanghun17@unist.ac.kr, kwonc@unist.ac.kr}}
\thanks{Digital Object Identifier (DOI): see top of this page.}%
}

\begin{document}

\markboth{IEEE Robotics and Automation Letters. Preprint Version. Accepted October, 2024}{LEE \MakeLowercase{\textit{et al.}}: Kernel-based Metrics Learning for Uncertain Opponent Vehicle Trajectory Prediction in Autonomous Racing}

\maketitle


\begin{abstract}
Autonomous racing confronts significant challenges in safely overtaking Opponent Vehicles (OVs) that exhibit uncertain trajectories, stemming from unknown driving policies. To address these challenges, this study proposes heterogeneous kernel metrics for Deep Kernel Learning (DKL), designed to robustly capture the diverse driving policies of OVs, and carry out precise trajectory predictions along with the associated uncertainties. A key virtue of the proposed kernel metrics lies in their ability to align similar driving policies and disjoin dissimilar ones in an unsupervised manner, given the observed interactions between the Ego Vehicle (EV) and OVs. The efficacy of the proposed method is substantiated through experimental studies on a 1/10th scale racecar platform, demonstrating improved prediction accuracy and thereby safely overtaking against OVs. Furthermore, our method is computationally efficient for onboard computing units, affirming its viability in fast-paced racing environments. The video and source code can be found at \url{https://github.com/HMCL-UNIST/OpponentPredictionWithKMDKL.git}.
\end{abstract}

\section{Introduction}
\IEEEPARstart{A}{utonomous} racing has emerged as a significant subfield of autonomous driving, attracting considerable interest and fostering competitions such as Roborace, Indy Autonomous Challenge, and F1TENTH \cite{betz2022autonomous}. One of the key challenges in autonomous racing lies in safely running against OVs and executing overtaking maneuvers. Various strategies have been developed to address these challenges \cite{wang2021game,zhu2023gaussian}, among which a commonly adopted approach involves predicting the future trajectory of OV and then planning the overtaking maneuver accordingly \cite{betz2022autonomous}.

Although previous studies have made strides in trajectory prediction for autonomous racing, they still face challenges in addressing diverse driving policies of OVs \cite{gulzar2021survey}. On the one hand, physics-based prediction methods are grounded in first-principles dynamics models. They are capable of anticipating momentary behavior but struggle to capture the long-term interactions between EV and OVs \cite{gulzar2021survey}. On the other hand, planning-based prediction methods infer the OVs' actions as solutions to optimization problems that account for the strategic objectives of the race and tactical interactions with the EV \cite{wang2021game}. However, these methods often demand intensive computation, which is critical for fast-paced racing environments. Moreover, the prediction performance heavily relies on the fidelity of the optimization problem, which necessitates encoding the OV's real objective \cite{gulzar2021survey}. Lastly, recent advancements in learning-based methods, which empirically examine driving data patterns, have shown promising results in trajectory prediction \cite{alahi2016social,varadarajan2022multipath++}. Nonetheless, data scarcity often hinders the implementation of learning-based methods within the autonomous racing regime. Moreover, when the encountered driving policy of the OV deviates from the training data, the resulting predictions fall short in planning safe maneuvers of the EV against OV \cite{betz2022autonomous}.

\begin{figure}[tpb]
    \centering
    \includegraphics[width=0.5\textwidth]{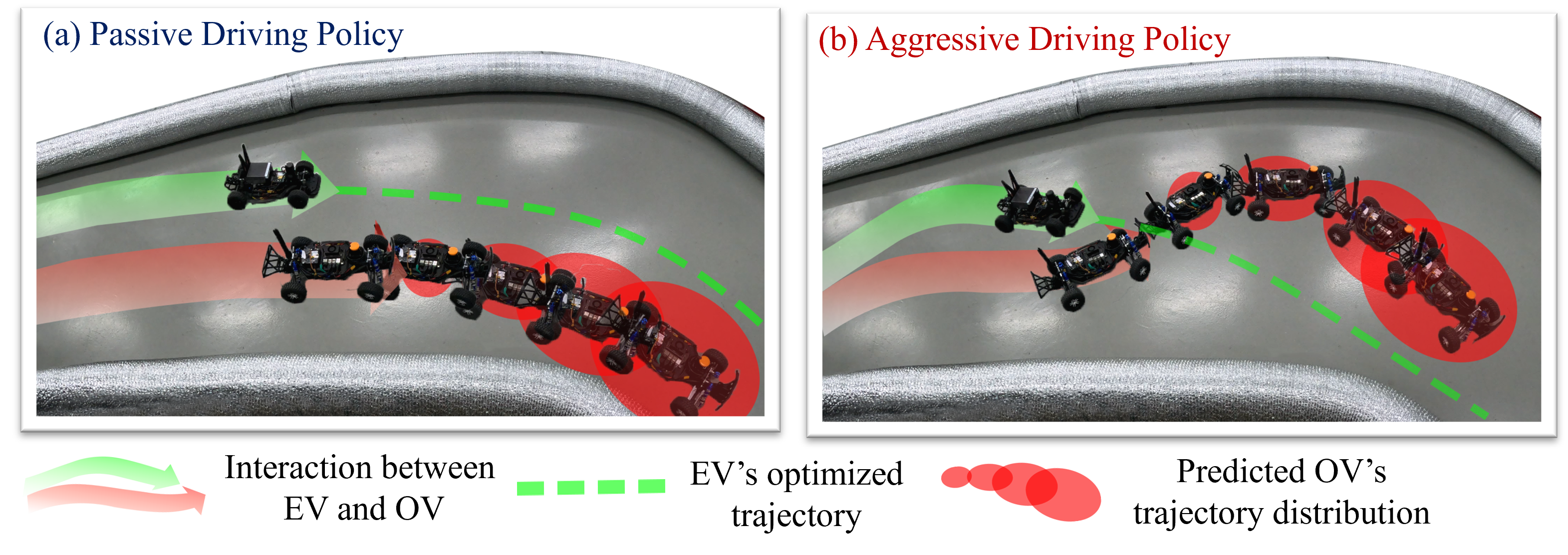}\caption{Overtaking scenarios where the EV maximizes its progress while taking into account the predicted trajectory distribution of the OV, which is based on the driving policy inferred from the observed EV-OV interaction. (a) OV exhibits non-blocking behavior, allowing the EV to navigate without interference; (b) OV actively tries to block the EV's progress.\vspace{-6mm}}\label{fig:main}
\end{figure}

This paper presents a learning-based method that can address the diverse driving policies of OVs, thereby accurately predicting the OV's trajectory and further assessing the uncertainty of prediction results (See Fig. \ref{fig:main}). First, the one-step OV's state prediction model is trained from the observed interaction between the EV and OV through DKL. In detail, a multi-scale Convolutional Neural Network (CNN) is employed to learn the interactions between the EV and OV, transforming the interaction patterns into latent representations of the OV's driving policy in an unsupervised manner. The represented OV's driving policy is then processed through a Gaussian Process (GP) model to predict the OV's state at the next time step. Building upon this process, we propose novel heterogeneous kernel-based metrics whereby DKL can more adeptly ascertain driving policy from complex interaction patterns. The proposed metrics are particularly devised to assess the similarities across different spaces, i.e., the latent space representing driving policies and the output space depicting the OV’s state prediction. Then, these metrics are optimized to render the latent space of DKL such that similar driving policies are clustered closely, while dissimilar ones are simultaneously spread apart. Such a strategic alignment improves the GP model's ability to capture the correlation between the driving policy and the state prediction. As a result, the DKL model not only enhances prediction performance but also addresses calibrated uncertainty concerning both trained and untrained driving policies.

Next, the one-step OV state prediction model is recursively executed to generate multi-step trajectory samples. These samples are utilized to construct the probabilistic OV trajectory distribution, predicting the state at each time step by its mean and variance. Subsequently, the predicted trajectory distribution information is integrated into the EV's Model Predictive Control (MPC) framework as dynamic obstacle constraints. This enables the EV to execute safe and agile overtaking maneuvers while considering the OV's uncertain predicted state distributions. To the best of the authors' knowledge, the proposed method for the first time attempts to apply kernel-based metrics learning to the realm of autonomous racing. Our method marks a notable advancement in OV trajectory prediction and surpasses the existing work in racing against OVs with diverse driving policies. The contributions of this paper are highlighted as follows: 
\begin{itemize}
    \item A novel kernel-based metrics learning framework for DKL is proposed, aimed at representing the diverse driving policies of OVs that are aligned with respect to their similarity.     
    \item A probabilistic OV trajectory prediction model is developed, improving prediction accuracy and uncertainty calibration based on the inferred OV's driving policy.
    \item An MPC-based planning for EV maneuver is seamlessly integrated with the trajectory prediction results, empowering the EV to safely overtake the OV.    
    \item Extensive real autonomous racing experiments with a 1/10th scale racecar demonstrate the effectiveness of the proposed algorithm, both quantitatively and qualitatively, compared to the existing algorithms.
\end{itemize}

\section{Related Work}
\subsection{Physics and planning-based trajectory predictions}
Physics-based trajectory prediction methods exploit the kinematics and dynamics of OVs, tailored by statistical filtering techniques \cite{gulzar2021survey}. However, these methods often face challenges in long-term forecasting due to their limited understanding of the interactions between the EV and OVs, implying a lack of consideration for the underlying driving policies of OVs. To account for the interactions, planning-based methods presume that the OVs logically decide their actions through optimization \cite{wang2021game}. These methods consider OVs as rational entities seeking to maximize their internal objectives. However, the accuracy of predictions hinges on the coherence between the presumed objective and the true objective of the OV. Furthermore, planning-based methods often demand heavy computations \cite{chen2022interactive}, presenting a formidable burden in fast-paced racing circumstances.

\subsection{Learning-based trajectory prediction}
Recent trajectory prediction research has primarily shifted towards learning-based methods that account for the complex interactions involving human social behavior \cite{alahi2016social} and multiple vehicles in urban settings \cite{varadarajan2022multipath++}. Nevertheless, their applicability degenerates when situated in autonomous racing, mostly due to deficient datasets in both quantity (limited number of racing scenes) and quality (lack of semantic cues) \cite{feng2024unitraj}. Moreover, the diversity in driving policies across different OVs further complicates learning the prediction model, requiring pertinent representations of driving policies.
Such a complication is further exacerbated by imbalanced datasets (e.g., overfitting on the dominant driving policies in the training data), making it difficult to generalize the prediction model. In response, methods such as meta-learning and representation learning have been explored to address the generalization of the prediction model against diverse driving policies \cite{liang2020simaug}. These methods employ distance metrics to capture complex data patterns and establish robust representations for similar/dissimilar patterns, improving prediction performance while reducing the risk of overfitting. While the choice of distance metrics, such as Euclidean, cosine, and kernel-based metrics, greatly impacts the representation model, determining the optimal metrics for measuring similarities remains an unresolved task to date \cite{zheng2023simts}. Lastly, the learning-based prediction model can be susceptible to out-of-training data distribution, meaning it is prone to yield unreliable prediction results when the encountered OV’s driving policy deviates from the training data \cite{gilles2022uncertainty}.

\subsection{Trajectory prediction with uncertainty}
Extensive investigation has been carried out to address uncertainty in trajectory prediction \cite{varadarajan2022multipath++}, among which GP has gained substantial attention due to its quantitative measure of uncertainty. Taking these advantages, GP has proven its effectiveness in cut-in behavior prediction \cite{yoon2021interaction}, and multi-step trajectory prediction of OV in autonomous racing \cite{zhu2023gaussian}. Typical GP describes the unknown function using stationary kernels, assuming that the properties of the function are constant across all input values. However, this often results in poor uncertainty calibration for non-stationary stochastic processes that exhibit intricate data patterns in practice \cite{wilson2016deep}. To handle the non-stationary aspect within GP, different techniques have been explored, such as input-dependent parametric kernels, Deep GP, and DKL \cite{wilson2016deep}. In particular, DKL leverages the non-linear mapping capability of neural networks to transform complex non-stationary input space into the latent feature space that can be effectively processed with stationary kernels in GP. However, training these complex kernels remains a significant challenge, as highlighted by recent studies \cite{ober2021promises}. Addressing this issue, this paper introduces a kernel-based metrics learning framework for DKL. This framework presents heterogeneous kernels to fine-tune the alignment of latent features transformed by the neural network, thereby adequately embedding the non-stationary properties into DKL.


\vspace{2mm}
\section{Problem Formulation}
\subsection{Vehicle dynamics}
The racing vehicles are modeled using a dynamic bicycle model \cite{liniger2015optimization}, where the state is represented by the vector $X=[p_{x}, p_{y}, \psi_{z}, v_{lon}, v_{lat}, w_{z}]^{\mathsmaller T}$. This vector consists of the vehicle's pose in Cartesian frame, $[p_{x}, p_{y}]^{\mathsmaller T}$, the heading angle, $\psi_{z}$, the longitudinal and lateral velocities, $[v_{lon},v_{lat}]^{\mathsmaller T}$, and the yaw rate, $w_{z}$. The input vector $u=[F_{xx}, \delta]^{\mathsmaller T}$ comprises the rear longitudinal tire force $F_{xx}$ and the front steering wheel angle $\delta$. To discretize the vehicle's nonlinear dynamics, we apply the $4^{th}$ order Runge-Kutta method with a sampling time step size $T_s$. The resulting discrete-time vehicle dynamics is compactly described as $X_{k+1} = f(X_{k}, u_{k})$, where the subscription $k$ indicates the time index. To examine the racing vehicle's motion, we interpret the vehicle's pose as Frenet-Serret formulas with respect to the racing track's centerline \cite{liniger2015optimization}. Let $\mathcal{C}:\mathbb{R}^{4} \mapsto \mathbb{R}^{4}$ denotes the invertible operator which projects the vehicle's pose in Cartesian frame to the curvilinear frame. Accordingly, the pose in curvilinear frame is denoted as $c = \mathcal{C}(p)$ where $p = [p_x, p_y, \psi_z, v_{lon}]^{\mathsmaller T}$ and $c = [s,e_{lat},e_{\psi}, v_{lon}]^{\mathsmaller T}$. Here, $s$ is the distance progress along the track's center line, $e_{lat}$ is the lateral offset from the center line, and $e_\psi$ is the angular deviation from the center line's tangent. The track's signed curvature at any distance $s\in[0,\tau]$ is represented as $\eta(s)$, where $\tau$ is the track length. To distinguish the state variables of EV and OV, we use superscripts in notations like $X^{ev}, c^{ev}, X^{ov}$, and $c^{ov}$.
\subsection{EV trajectory planning problem for racing}
The EV trajectory planning problem is formulated as Model Predictive Contouring Control (MPCC), which allows the EV to plan its trajectory over $N$ steps ahead at time step $k$, defined as 
$\mathbf{\hat{X}}^{ev}_{k:k+N} = \{\hat{X}_{k}^{ev},\ldots, \hat{X}_{k+N}^{ev}\}$, with respect to the projected distance progress $\bar{s}_{t}$ on the centerline \cite{liniger2015optimization}. Further, the constraints of MPCC related to dynamic obstacles are augmented to incorporate the predicted OV trajectory. Given the predicted trajectory of the OV over $N$ steps ahead at time step $k$, i.e., $\{\hat{X}_{k}^{ov}, \ldots, \hat{X}_{k+N}^{ov}\}$, the MPCC problem for the EV is formulated as follows:
\begin{subequations}    
\begin{multline}
\min_{\mathbf{u},\mathbf{v}} \sum_{t=0}^{N-1} ||e_{c}(\hat{X}^{ev}_t,\bar{s}_t)||_{q_{c}}^{2} +||u_t||^{2}_{R_u} \\ 
+ ||u_{t}-u_{t-1}||^{2}_{R_d} - q_s\bar{v}_N \label{eq:obj}
\end{multline}\label{eq:MPCform} 
     \vspace{-6mm} 
\begin{align}
s.t. \ &  \hat{X}^{ev}_{0} = X^{ev}_{k}, \ \bar{s}_0 = s^{ev}_k, \ u_{-1} = u^{ev}_{k-1}, \hspace{-20mm} \label{eq:mpccinit}\\ 
&\hat{X}^{ev}_{t+1} = f(\hat{X}^{ev}_t,u_t), \ &t= 0,....,N-1  \label{eq:mpcdyn}\\
&\bar{s}_{t+1} = \bar{s}_t + T_s \bar{v}_{t}, \ &t= 0,...,N-1 \label{eq:mpccapx}\\
&\hat{X}^{ev}_t \in X_{\text{track}}, \  u_t \in \mathbf{U}, \ &t= 0,...,N \label{eq:mpccconst}\\
&h(\hat{X}^{ev}_t,\hat{X}^{ov}_{k+t}) \leq 0, \ &t= 0,...,N \label{eq:mpccobs} \end{align}
\end{subequations}
where $\mathbf{u}=\{u_0,...,u_{N-1}\}$, $\mathbf{v}=\{\bar{v}_0, \ldots, \bar{v}_{N-1}\}$. $\bar{v}_t$ represents the progression rate, and $e_{c}$ measures the approximated projection errors between $\hat{X}_{t}^{ev}$ and the reference centerline at $\bar{s}_{t}$ \cite{liniger2015optimization}. The objective cost \eqref{eq:obj} conceives of maximizing the race progress while penalizing the centerline deviation according to weighting parameters, $q_c, \ q_s > 0$. To ensure a smooth control profile, cost minimizes the magnitude and rate of the input with the weights $R_u$ and $ R_d \succ 0$, respectively. The dynamics of the EV is governed by Eqs. \eqref{eq:mpcdyn} and \eqref{eq:mpccapx}. In \eqref{eq:mpccconst}, $X_{track} = \{ X| -W_{track}/2 \leq e_{lat} \leq W_{track}/2 \}$ and $\mathbf{U}$ represent the constraints imposed by the track boundary and the control input limits, where $W_{track}$ is the track width. The constraint in Eq. \eqref{eq:mpccobs} ensures collision avoidance, given that the predicted OV state forms a probability distribution, such as Gaussian distribution $\hat{X}_{k+t}^{ov} \sim \mathcal{N}(\mu(\hat{X}_{k+t}^{ov}), \sigma^2(\hat{X}_{k+t}^{ov}))$. In constructing Eq. \eqref{eq:mpccobs}, the EV's safety margin is drawn as a collection of four circles, while an ellipse represents the OV's margin. To accommodate the uncertainty encoded in the predicted state distribution $\hat{X}_{k+t}^{ov}$, the size of the OV's ellipse is dynamically adjusted in proportion to its variance, $\sigma^{2}(\hat{X}_{k+t}^{ov})$. Detailed derivation for the ellipse size can be found in Section IV of \cite{zhu2023gaussian}, which is omitted here for brevity. Notably, the ellipsoidal safety margin is widely adopted due to its effectiveness in capturing Gaussian-distributed uncertainties in the predicted OV state and its computational efficiency in solving nonlinear MPC problems \cite{reiter2023frenet}. 

\begin{figure*}[ht]
\centering
\includegraphics[width=0.95\linewidth]{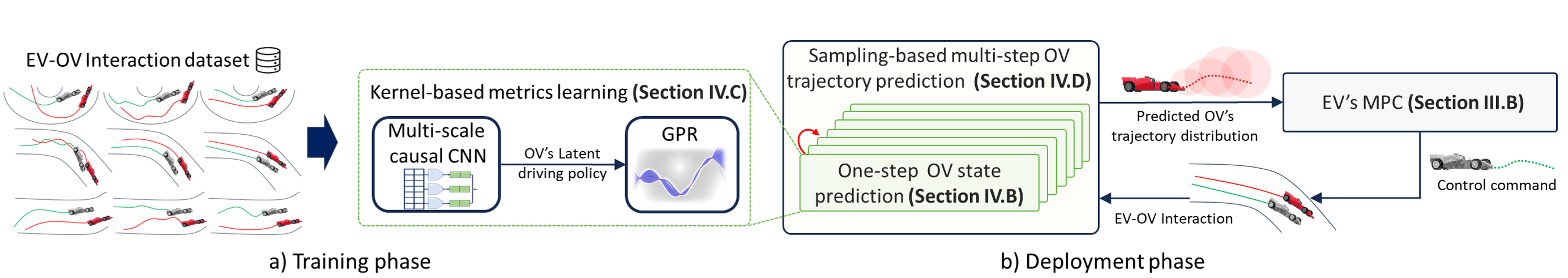}
\caption{The architecture of the proposed algorithm: (a) the training process of the one-step state prediction model, which utilizes the proposed kernel-based metrics learning for DKL; and (b) an autonomous racing algorithm for EV, which plans its maneuver based on the predicted trajectory distribution of OV.\vspace{-3mm}}
\label{fig:main_alg_row}
\end{figure*}

\subsection{GP and DKL for OV trajectory prediction}
To predict the OV trajectory which is implicated in the patterns of both EV and OV states, we employ a GP Regression (GPR) model to be trained using the EV-OV interaction dataset. GPR is a non-parametric Bayesian regression that embeds a GP, characterized by a mean function $m(\cdot)$ and a covariance function $\kappa(\cdot,\cdot)$, commonly referred to as a kernel. Typically assuming a zero-mean GP prior for simplicity, the posterior distribution of GPR is given by $\hat{f}(x)\sim \mathcal{N}(\bar{m}(x),\bar{\sigma}(x))$ where $\bar{m}(x)$ and $ \bar{\sigma}(x)$ represent the posterior mean and covariance functions calculated from the training data, respectively.

In this paper, we adopt the Mat$\acute{e}$rn kernel, $\kappa(x,x')$, which has proven to be effective in the autonomous driving domain \cite{yoon2021interaction,zhu2023gaussian}. 
Mat$\acute{e}$rn kernel, however, is a stationary kernel that may be insufficient to capture the complex patterns in the trajectories of different OVs with diverse driving policies. To address this, DKL applies a non-linear transformation to the input space and utilizes the same stationary kernels for the transformed space, namely feature space. Accordingly, the kernel in DKL can be reformulated as:
\begin{equation*} 
\bar{\kappa}(x,x') = \kappa(\mathbf{\Phi}(x), \mathbf{\Phi}(x'))
\end{equation*}
where $\mathbf{\Phi}: \mathbb{R}^{n} \rightarrow \mathbb{R}^{n'}$ denotes the transformation function designed as a neural network, mapping inputs to feature space. However, learning the optimal design of $\mathbf{\Phi}$ in an unsupervised manner is challenging due to complex EV-OV interactions that yield diverse OV driving policies.




\section{Algorithm Development}
\subsection{Algorithm overview}
In this section, we introduce the main algorithm to predict the trajectory distributions of OVs. First, the one-step OV state prediction model is developed using DKL, which is trained on the EV-OV interaction dataset. Then, we introduce a DKL training framework where novel kernel metrics are employed to address the diversity of OV driving policies. Based on the trained DKL, a sampling-based method is applied to the one-step prediction model to yield a multi-step predicted trajectory distribution. The overall training and deployment phases of the proposed OV trajectory prediction are outlined in Fig. \ref{fig:main_alg_row}.


\subsection{One-step OV state prediction using DKL}
To model the one-step state prediction of the OV, we employ DKL, whose input is the past trajectories of both EV and OV, signifying their interaction patterns over a high-dimensional space. The input for the DKL at time step $k$ is formulated as a sequence $\bold{z}_{k-M:k}=[ \bold{z}_{k-M}^{\mathsmaller T}, \ldots, \bold{z}_{k}^{\mathsmaller T}]^{\mathsmaller T}\in\mathbb{R}^{10\times M}$, where $M$ denotes the sequence length that records the past interactions, including the current states of EV and OV. Specifically, the vector $\bold{z}_{k}$ includes the state information of both vehicles along with track geometry, defined as:
\begin{equation*}
    \bold{z}_{k} = [(s_{k}^{ov}-s_{k}^{ev}), e_{lat,k}^{ov}, e_{\psi,k}^{ov}, v_{lon,k}^{ov},e_{lat,k}^{ev}, e_{\psi,k}^{ev}, v_{lon,k}^{ev}, \tilde{\eta}_{k}^{ov}]^{\mathsmaller T}
\end{equation*}
where $\tilde{\eta}^{ov}_{k} = [\eta(s^{ov}_{k}), \eta(s^{ov}_{k}+\zeta),\eta(s^{ov}_{k}+2\zeta)]^{\mathsmaller T}\in\mathbb{R}^{3}$ represents the vector of track curvatures at look-ahead distances $s_k^{ov}+i\zeta,\ i=0,1,2$ with some offset $\zeta>0$. Notably, the vector $\mathbf{z}_{k}$ is crafted to capture the interaction between the EV and OV, specifically contextualized within the racing track geometry. This facilitates a more accurate description of interaction patterns relative to the track layout, rather than relying solely on the absolute global pose.

The DKL transforms the input into the feature space, representing the OV's latent driving policy as follows: 
\begin{equation}\label{eq:latent}
\hat{\phi}_k^{ov} := \mathbf{\Phi}(\mathbf{z}_{k-M:k})
\end{equation} 
where $\mathbf{\Phi}: \mathbb{R}^{10\times M} \rightarrow \mathbb{R}^{L}$ is a nonlinear mapping function that encodes the interaction history between EV and OV over the past $M$ steps. By this definition, $\mathbf{\Phi}$ is designed to map different EV-OV interaction patterns to OV driving policies. In this paper, we implement $\mathbf{\Phi}$ as a neural network model trained on racing data against various OVs. Consequently, the resulting $\hat{\phi}_{k}^{ov}$ exhibits ``diverse driving policies" that arise from the different interaction patterns of the OVs. In pursuit of extracting an informative policy representation from the interaction data, we adopt a multi-scale encoder architecture as detailed in \cite{zheng2023simts}. This architecture employs multi-filters with varying kernel sizes to capture both global and local interaction patterns, utilizing 1D causal convolution layers to preserve only the causal relationships in the data sequence. Subsequent to the multi-filtering, the final layers average the outputs of different kernel sizes, resulting in a comprehensive representation of the interaction patterns.

Subsequently, the driving policy $\hat{\phi}_k^{ov}$ inputs into GPR to predict the uncertain state propagation of the OV at the next time step, expressed by the following output vector:
\begin{equation*}
\begin{split}
    y_k = [ &  
    s_{k+1}^{ov} - s_{k}^{ov},
    e_{lat,k+1}^{ov} - e_{lat,k}^{ov}, \\ 
    & e_{\psi,k+1}^{ov} - e_{\psi,k}^{ov},
    v_{lon,k+1}^{ov} - v_{lon,k}^{ov}
    ]^{\mathsmaller T}
\end{split}    
\end{equation*}
This one-step state difference $y_k\in\mathbb{R}^{4}$ is modeled as unknown stochastic processes $g$ as follows:
\begin{equation*}
 y_k =[y^{(1)}_k,y^{(2)}_k,y^{(3)}_k,y^{(4)}_k]^{\mathsmaller T}= g(\mathbf{\Phi}(\mathbf{z}_{k-M:k}))  
\end{equation*}
Without loss of generality, we assume the individual output elements $y^{(1)},y^{(2)},y^{(3)}$, and $y^{(4)}$ are independent of each other. Then, the GPR produces a probability distribution for each output element $y^{(n)}$ as follows:
\begin{equation*}
\begin{split}   
y^{(n)}& = g^{(n)}(\mathbf{\Phi}(\mathbf{z}_{k-M:k})) \\
&\sim \mathcal{N}(\mu^{(n)}(\mathbf{\Phi}(\mathbf{z}_{k-M:k})), (\sigma^{(n)}(\mathbf{\Phi}(\mathbf{z}_{k-M:k})))^2)
\end{split}
\end{equation*}
where $\mu^{(n)}$ and $(\sigma^{(n)})^2,\ n=1,\ldots, 4$, represent the posterior means and variances, respectively. It is noted that the obtained variances of the GPR capture the epistemic uncertainty, primarily emerging from the diversity of OVs' driving policies, as well as aleatoric uncertainty stemming from measurement noise.

\subsection{Kernel-based metrics learning for DKL}
In this section, a novel training framework for DKL is introduced. The DKL is typically optimized in regard to the negative log marginal likelihood loss, $\mathcal{L}_{mll}$. However, solely optimizing $\mathcal{L}_{mll}$ is prone to overfitting and often yields suboptimal performance during testing \cite{ober2021promises}. To address this challenge, we incorporate additional regularization to better align the latent driving policies, $\hat{\phi}_k^{ov}$, computed by the neural network transformation, $\mathbf{\Phi}$. Central to our approach is the strategic clustering of similar latent driving policies while distancing dissimilar ones, based on the similarities in the output space, i.e., OV's state differences.

\begin{figure}[htpb]
\centering
\includegraphics[width=0.45\textwidth]{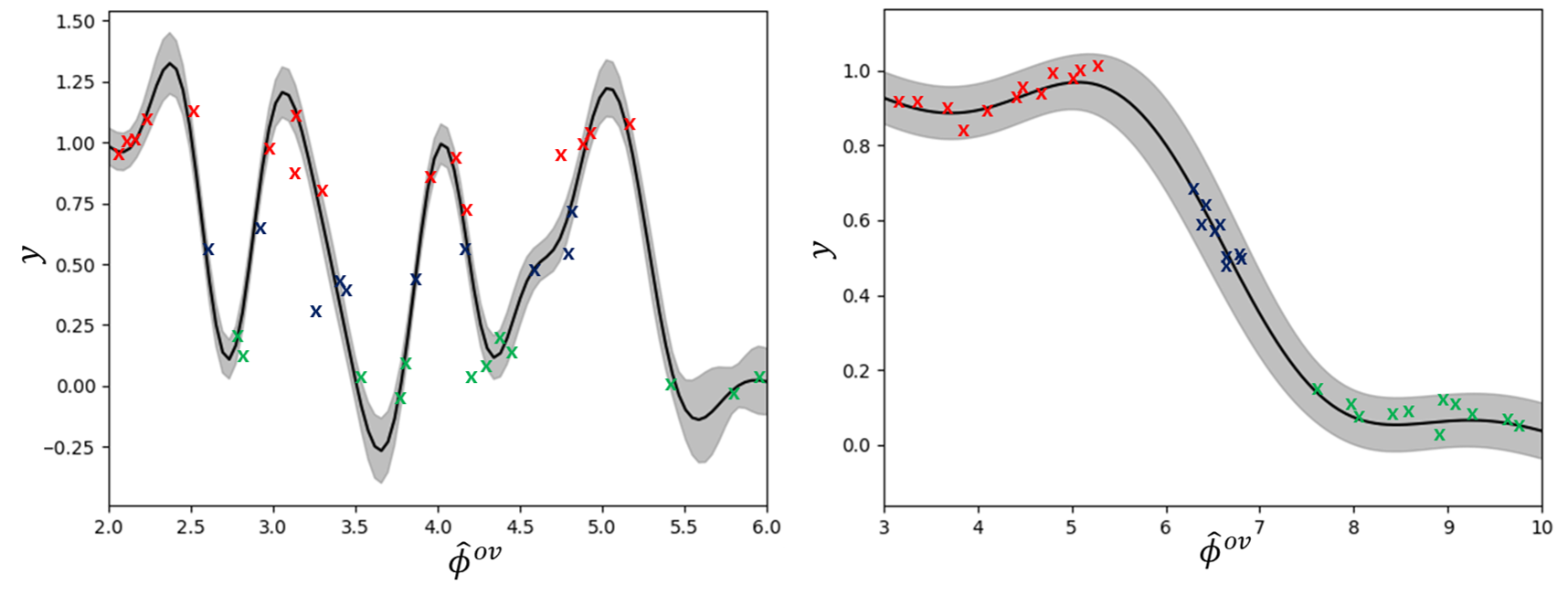}
\caption{Impact of latent space alignment on DKL prediction model: (Left) Without latent space alignment, (Right) With latent space alignment achieved by clustering similar features and dispersing dissimilar ones.\vspace{-2mm}}\label{fig:hetkernel}
\end{figure}

Fig. \ref{fig:hetkernel} illustrates the impact of alignment in latent driving policies on the performance of GPR given the same stochastic unknown processes. In the left subfigure, the input and output arrangement is disorganized, leading to increased susceptibility to overfitting and difficulties in training a GPR. Such a poor alignment eventually degrades the prediction model. In contrast, the right subfigure demonstrates a strategically aligned arrangement of latent driving policies, mirroring similarities observed in the output space. This, in turn, ensures that the driving policy $\phi^{ov}_{k}$ is robustly represented, enabling the trained OV state propagation model, $g$, to accurately predict the output $y_{k}$ from  $\phi^{ov}_{k}$, even when $\phi^{ov}_{k}$ is influenced by variations or uncertainties in the input $\mathbf{z}_{k-M:k}$. As a result, this alignment reduces the risk of overfitting and improves robustness by minimizing sensitivity to input variations, thereby enhancing overall prediction performance.


To align the latent driving policies, we introduce a distance-matching loss term denoted as $\mathcal{L}_{dist}$ to minimize the dissimilarity between the distances in the pair of latent features and the distances in the pair of outputs. This loss for a mini-batch $\mathcal{D}= \{\hat{\phi}_i^{ov}, y_{i}\}_{i\in[1:D]}$ can be written as:
\begin{equation}\label{eq:dist}
    \mathcal{L}_{dist} = \frac{1}{D}\sum_{i\neq j}^{D}||\kappa_{\mathbf{\Phi}}(\hat{\phi}_i^{ov}, \hat{\phi}^{ov}_j) - \kappa_{y}(y_i, y_j) ||_{2}
\end{equation}
where $D$ is the size of the mini-batch. $\kappa_{\mathbf{\Phi}}$ and $\kappa_{y}$ are Mat$\acute{e}$rn kernels, the same type of kernel used in GPR. The prediction model regularized by Eq. \eqref{eq:dist} enforces the data to be aligned in the relevant distance.

On the other hand, the extent of alignment is strongly subject to the hyperparameters of $\kappa_{\mathbf{\Phi}}$ and $\kappa_{y}$, i.e., lengthscales $l_{\mathbf{\Phi}}$ and $l_{y}$, respectively. These hyperparameters directly influence the representation of data, thereby affecting the alignment of latent features. For instance, a smaller lengthscale in the kernel increases its sensitivity to small differences between data, causing them to be treated as significantly dissimilar. Conversely, a larger lengthscale reduces this sensitivity, allowing for a broader margin to identify similarities among distanced data. 
\begin{figure}[htpb]
\centering\
\includegraphics[width=0.48\textwidth]{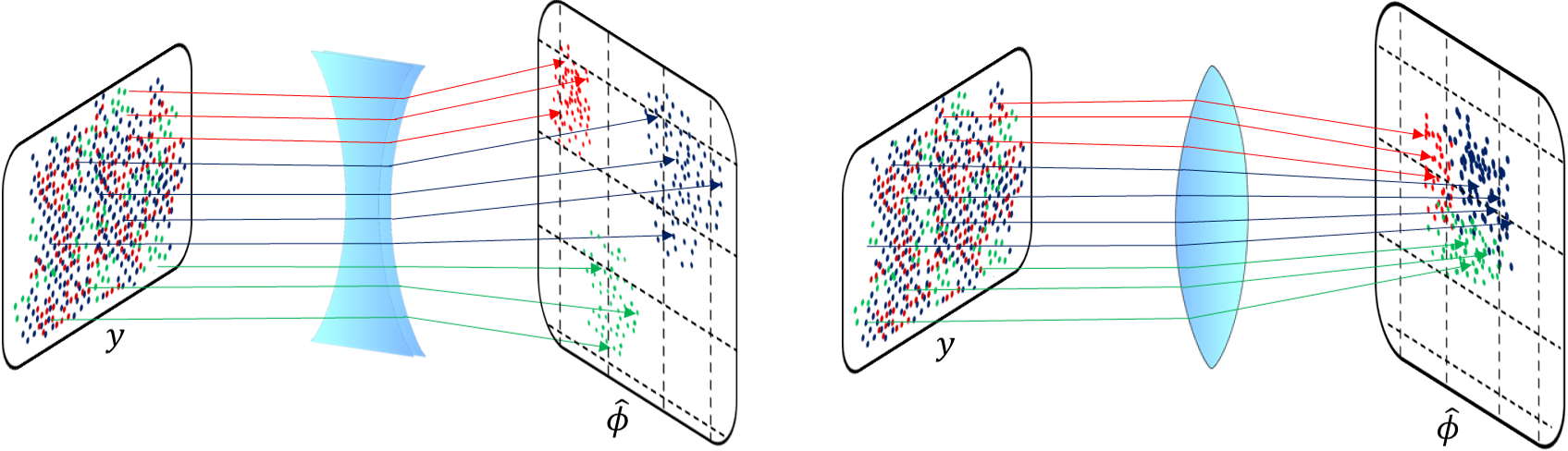}
\caption{Alignment in latent space with varying kernel lengthscales: (Left) $l_{y} < l_{\mathbf{\Phi}}$, (Right) $l_{y} > l_{\mathbf{\Phi}}$.\vspace{-2.5mm}}\label{fig:lens}
\end{figure}
To find the best lengthscales for enhanced clustering of similar data and dispersion of dissimilar ones in latent space, we adjust the kernel sensitivity: decreasing it in the latent space, $\kappa_{\mathbf{\Phi}}$, to widen the distribution of dissimilar data, while increasing it in the output space, $\kappa_{y}$, for precise capture of data differences. This dual regularization is facilitated by introducing a sensitivity loss as follows:
\begin{equation}\label{eq:sense}
 \mathcal{L}_{sense} = \ln(\frac{l_{y}}{l_{\mathbf{\Phi}}})+ \textrm{max}(0, w_{1}(\sigma_{sense}(\hat{\phi}^{ov}) - \alpha))
\end{equation}
where $w_{1}$ and $\alpha$ denote the predefined weight and the threshold, respectively. The second term in Eq. \eqref{eq:sense} prevents excessive divergence in the latent space, where the standard deviation of the latent features of the batch data, i.e., $\{\hat{\phi}_{i}^{ov}|i\in 1,\ldots,D\}$, is denoted as $\sigma_{sense}(\hat{\phi}^{ov})$. Fig. \ref{fig:lens} graphically visualizes the effectiveness of the proposed regularization by Eq. \eqref{eq:sense}. Here, the left-hand side illustrates the preferred scenario in which the different latent driving policies are more widely dispersed across various data. 

Incorporating all the aforementioned loss terms, DKL is trained to optimize both the neural network and the GP's hyperparameters by minimizing the following composite loss:
\begin{equation}\label{eq:dkl}
    \mathcal{L}_{DKL} = \mathcal{L}_{mll}  + w_{2} \mathcal{L}_{dist} + w_{3} \mathcal{L}_{sense} 
\end{equation}
where $w_{2}$ and $w_{3}$ are scaling constants. In a nutshell, our distinct novelty lies in the inclusion of both $\mathcal{L}_{dist}$ and $\mathcal{L}_{sense}$ in DKL. While $\mathcal{L}_{mll}$ addresses the nominal DKL model for prediction, $\mathcal{L}_{dist}$ is essential for appropriately aligning the latent driving policies and $\mathcal{L}_{sense}$ is for fine-tuning of kernel sensitivity. Without $\mathcal{L}_{sense}$, kernel-based metrics for both output and latent space may exhibit reduced sensitivities, undermining the clustering efficacy intended by $\mathcal{L}_{dist}$. It is to be noted that $\mathcal{L}_{dist}$ and $\mathcal{L}_{sense}$ are mutually complementing each other to achieve the latent feature alignment. Therefore, their interdependency obviates the need for an ablation study of each loss term individually.

\vspace{-1mm}


\subsection{Sampling-based multi-step OV trajectory prediction}
The one-step OV state prediction model, trained by the proposed DKL framework, is used to predict the OV trajectory over the $N$ step horizon. The basic idea of this multi-step prediction is consecutively executing one-step predictions over future time steps, considering not only the past interaction history but also future interaction. This requires the predicted EV trajectory, $\mathbf{\hat{X}}^{ev}_{k:k+N-1}$, which can be obtained from the open-loop solution to \eqref{eq:MPCform} for the EV motion planning at $k-1$ time step.

Analytically computing the solution to the multi-step prediction model is intractable in general \cite{zhu2023gaussian}. To address this, we utilize a sampling-based approach with $Q$ samples to estimate the multi-step state propagation of the OV. Subsequently, we calculate the statistics of these samples to derive the nominal trajectory prediction along with its variances. Specifically, for the $i^{th}$ sample, starting at time step $k$, we sample OV states at time step $k+1$ as $c^{ov,i}_{k+1}$ using one-step prediction. Given the predicted states of the EV and OV, we construct $\mathbf{z}^{i}_{k+1}$ along with the track information. This vector is used to update the input to the one-step prediction model at the next time step, $\mathbf{z}^{i}_{k-M+1:k+1}$. We then continue to roll out the consecutive steps for multi-step prediction, repeating over $N$ steps, resulting in the set of multi-step samples, $\{c^{ov,i}_{k+t}|\ i=1,\ldots, Q, t=1,\ldots,N\}$. Correspondingly, the predicted trajectory distribution at each future time step, $k+t$, is defined by its mean, $c_{k+t}^{ov}$, and covariance, $\Sigma_{k+t}^{ov}$, calculated as follows:
\begin{equation}\label{eq:mean_cov}
\begin{split}
c_{k+t}^{ov} &= \frac{1}{Q}\Sigma_{i=1}^{Q}c_{k+t}^{ov,i} \\
\Sigma_{k+t}^{ov} &= \frac{1}{Q-1}\Sigma_{i=1}^{Q} (c_{k+t}^{ov,i}-c_{k+t}^{ov}) (c_{k+t}^{ov,i}-c_{k+t}^{ov})^{\mathsmaller T}
\end{split}
\end{equation}
Lastly, the prediction outcomes are utilized to define collision avoidance constraints, as formulated in Eq. \eqref{eq:mpccobs}, ensuring safe and agile racing maneuvers of EV. The detailed procedure is described in Algorithm \ref{alg:algorithm}.

\begin{algorithm}[th]
\begin{algorithmic}
\State 
\textbf{Set}: Prediction horizon length $N$ and the number of samples $Q$.
\\ 
\textbf{Input} : Current EV and OV states, $c^{ev}_{k}$ and $c^{ov}_{k}$; EV's predicted states, $\mathbf{\hat{X}}^{ev}_{k:k+N-1}$; interaction history between EV and OV, $\mathbf{z}_{k-M:k}$.
\end{algorithmic}
\begin{algorithmic}[1]
    \State $c_k^{ov,i} = c_{k}^{ov}$, $\mathbf{z}_{k-M:k}^{i} = \mathbf{z}_{k-M:k}, \forall i= 1,\ldots, Q$
    \For{$ i =1,\ldots,Q$}
    \For{$ t =0,\ldots,N-1$}        
        \State Compute $\hat{\phi}_{k+t}^{ov,i}$ in Eq. \eqref{eq:latent} given  $\mathbf{z}_{k-M+t:k+t}^{i}$
        \State Sample $y^{i}_{k+t}\sim \mathcal{N}(\mu(\hat{\phi}_{k+t}^{ov,i}), \sigma^2(\hat{\phi}_{k+t}^{ov,i}))$ with GP
      \State $c^{ov,i}_{k+t+1}$ = $c^{ov,i}_{k+t} + y^{i}_{k+t}$
      \State \textbf{if} $t < N-1$ \ \textbf{then}
            \State \ \ Update $\mathbf{z}_{k-M+t+1:k+t+1}^{i}$ given $\hat{X}^{ev}_{k+t+1}$, $c^{ov}_{k+t+1}$
        \EndFor
    \EndFor
\For{$ t =1,\ldots,N$}  
    \State Compute $c_{k+t}^{ov}$ and $\Sigma_{k+t}^{ov}$ by Eq. \eqref{eq:mean_cov}
\EndFor 
\State \textbf{Output} : $c_{k+t}^{ov}, \Sigma_{k+t}^{ov} \ \ \forall t=1,..,N$ 
\vspace{-1mm}\end{algorithmic} 
\caption{Multi-step trajectory prediction with DKL}
\label{alg:algorithm}
\end{algorithm}

\vspace{-3mm}

\section{Hardware Experiments}
\subsection{Experiment setup}
\begin{figure}[htpb]
\centering\
\includegraphics[width=0.38\textwidth]{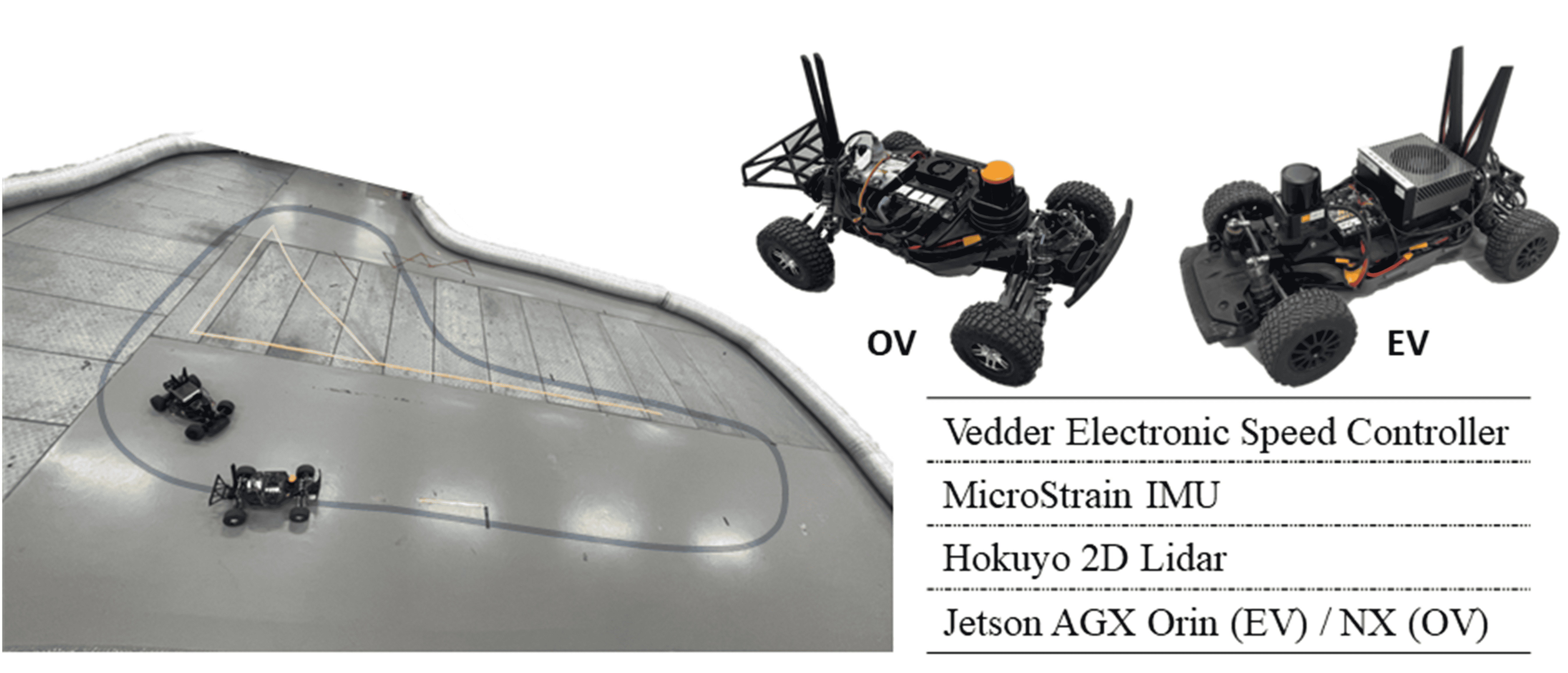}
\caption{1/10th scale racing experiments setup. L-shape racing track and the vehicles' specifications.\vspace{-6mm}}\label{fig:hardware}
\end{figure}

The EV and OV are built on the 1/10th scale racecar shown in Fig. \ref{fig:hardware}. To assure reliable real-time execution, the EV utilizes a Jetson AGX Orin, while the OV operates efficiently with a Jetson Orin NX. In terms of software configuration, both vehicles operate with ROS Noetic. They utilize the Cartographer \cite{hess2016real} to perform local pose estimation within a shared map common to both the EV and OV. The state estimation involves a customized estimator based on GTSAM \cite{gtsam}. Instead of detecting and tracking the states of other vehicles using sensor measurements, the EV and OV exchange their own state estimates via ROS topics. Both the EV and OV solve optimal control problems using the FORCESPRO Nonlinear Interior-Point solver \cite{FORCESNLP} through code generation. Additionally, DKL is implemented within the PyTorch framework, incorporating variational GPR from GPyTorch \cite{gardner2018gpytorch}. Control operations are executed at 20Hz, while multi-step trajectory predictions are performed at 10Hz. 

\subsection{Diverse driving policies of opponent vehicles}
The baseline driving policy for both the EV and OV is established by the MPCC formulation in \eqref{eq:MPCform} with a planning horizon of $N=12$ and the sample time of $T_s=0.1s$. To facilitate overtaking maneuvers and ensure close interaction between vehicles, longitudinal speed constraints of $1.9$m/s for the EV and $1.6$m/s for the OV are imposed, respectively. 

OV driving policies are designed to interact with the EV in two different manners: through direct blocking maneuvers (i.e., aggressive driving policy) or by focusing solely on minimal lap time without blocking (i.e., passive driving policy). The implementation of the OV's aggressive driving policy, as outlined in \cite{zhu2023gaussian}, actively obstructs the EV by minimizing the lateral distance between the OV and EV. This is achieved by including the blocking weights in the cost function \eqref{eq:obj}. Conversely, the passive driving policy is implemented by omitting the collision avoidance constraints in Eq. \eqref{eq:mpccobs} and not incorporating the blocking weights, thus disregarding any interactive maneuvers with the EV. 

To train the prediction model, two datasets, $\mathcal{D}_{aggr}$ and $\mathcal{D}_{pass}$, are collected to capture the interaction with OVs under these two different driving policies: aggressive and passive, respectively. For our one-step OV state prediction model, we set the dimension of the latent driving policy to $L=11$ and the observed interaction sequence length to $M=10$. The weights in Eqs. \eqref{eq:sense} and \eqref{eq:dkl} are tuned as $w_{1}=2$, $w_{2}=1$, and $w_{3}=0.05$. The training, validation, and test datasets are constructed using samples that are randomly pooled from $\mathcal{D}_{aggr}$ and $\mathcal{D}_{pass}$ without any annotations. Finally, for the multi-step prediction model, we set the number of samples in generating the predicted trajectory distribution to $Q=25$. 



\begin{figure*}[ht]
\centering
\includegraphics[width=0.92\linewidth]{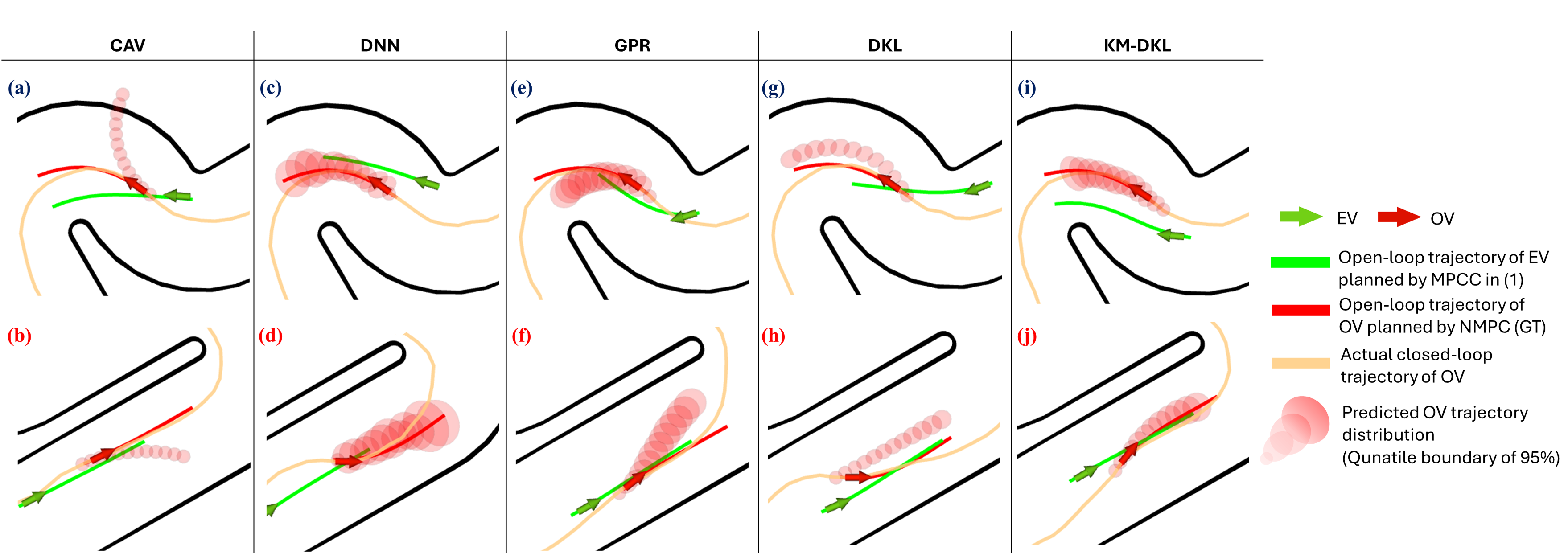}
\caption{Snapshots of EV racing maneuvers against OV with passive (a,c,e,g,i) and aggressive (b,d,f,h,j) driving policies. Black solid lines represent the track boundaries. A comprehensive video demonstration of the hardware experiments is accessible at \url{https://github.com/HMCL-UNIST/OpponentPredictionWithKMDKL.git}.\vspace{-2mm}}
\label{fig:single}
\end{figure*}

\begin{table*}[t]
\centering
\caption{Racing result statistics against OVs with diverse driving policies.}\vspace{-3mm}
\label{t2}
\begin{tabular}{c||c|c|c|c|c|c}
\noalign{\smallskip}\noalign{\smallskip}\hline
& NMPC(GT) & CAV & DNN \cite{varadarajan2022multipath++} & GPR \cite{zhu2023gaussian} & DKL & KM-DKL(proposed)  \\
\hline
Longitudinal MSE (mean $\pm$ std)  & \textbf{0.178}\ $\pm$\ 0.11 & 0.746\ $\pm$\ 0.307 & 0.839\ $\pm$\ 0.12 & 0.290\ $\pm$\ 0.15 & 0.256\ $\pm$\ 0.193 & 0.244\ $\pm$\ \textbf{0.09}  \\
Lateral MSE (mean $\pm$ std) & 0.217\ $\pm$\ 0.14  & 0.584\ $\pm$\ 0.35 & 0.377\ $\pm$\ 0.15 & 0.219\ $\pm$\ 0.14 & 0.261\ $\pm$\ 0.17 & \textbf{0.139}\ $\pm$\ \textbf{0.08}  \\
\hline
Overtaking rate per race & 0.85 & 0.55  &0.65 & 0.35 & 0.65 & \textbf{0.95}  \\
Collision rate per race (minor)  & \textbf{0.30} & 0.45 & 0.85 & 0.40 & 0.50 & \textbf{0.30}  \\
Collision rate per race (major)  & 0.10  & 0.30  & 0.25 & 0.55 & 0.30 & \textbf{0.00}  \\
Average computation time (ms)  & 12  & \textbf{6} & 39 & 84 & 97 & 98 \\
\hline
\end{tabular}\label{tb:race}
\vspace{-5mm}
\end{table*}

\subsection{Racing results and discussion}  
Extensive experiments are carried out to assess the prediction performance of the proposed method and further validate its effectiveness in head-to-head racing, particularly in terms of overtaking maneuvers. For comparative analysis, we evaluate our multi-step trajectory prediction method versus five distinct baseline trajectory predictors. The first baseline, referred to as NMPC (a.k.a Ground Truth, GT), utilizes a planning-based nonlinear MPC predictor. This method employs the open-loop solution to the OV's MPCC problem, serving as the ground truth prediction from the OV's perspective. The second baseline, denoted as Constant Angular Velocity (CAV), is a physics-based prediction method that propagates the OV's current state under constant linear and angular velocity assumptions. The third baseline, DNN, adapted from \cite{varadarajan2022multipath++}, is a deep learning-based method originally developed for urban environments. The input is modified to include EV and OV state histories and track information, i.e., $\mathbf{z}_{k-M:k}$. Additionally, the output provides the mean and variances of the multi-step predicted state propagation, which are matched with the output of Algorithm 1 for a fair comparison with other baselines. The fourth baseline, denoted as GPR, adapted from \cite{zhu2023gaussian}, is a learning-based method that employs GPR with a stationary Mat$\acute{e}$rn kernel to predict the multi-step trajectory together with its associated uncertainties. This method utilizes the current driving scene information only, i.e., $z_{k}$. Consequently, it faces limitations in capturing the interaction patterns between the EV and OV that are influenced by OV's driving policy. The fifth baseline denoted as DKL, represents an ablation version of our proposed method. It utilizes the same architecture as the one-step prediction model but omits the proposed kernel-based metrics learning. This is more or less nominal DKL that solely aims at maximizing the log marginal likelihood of observations, i.e., $\mathcal{L}_{mll}$. Lastly, our proposed method, incorporating a novel kernel-based metrics learning as in Eq. \eqref{eq:dkl}, is referred to as KM-DKL. Unlike the other methods that provide prediction results as trajectory distributions, CAV and NMPC (GT) predictors provide deterministic trajectories, lacking a probabilistic measure of prediction confidence. For the sake of fair comparison concerning uncertainty quantification ability, we set a fixed circular boundary to their prediction results.
\\ \indent The experiment involves twenty races for each baseline method. The OV is placed in various starting positions, which remain the same for all predictors. Each race concludes once the EV completes three laps or if a major collision occurs that drastically changes the course of both cars, making it impossible to continue the race. For each race, we set the OV's driving policy to either passive or aggressive, resulting in ten races for each policy. The prediction performance is then evaluated by the mean square errors of longitudinal and lateral poses between predicted and actual trajectories at the last prediction step when the EV is in the proximity to the OV, i.e., $0\leq s^{ov}-s^{ev}\leq2$. 
\\ \indent Notably, given the result statistics detailed in Table. \ref{t2} and the prediction snapshots in Fig. \ref{fig:single}, our method surpasses all the baselines in lateral prediction accuracy, with NMPC(GT) achieving the lowest longitudinal mean error. Even though NMPC(GT) anticipates the ground truth open-loop solution from OV's perspective, it fails to capture the multi-step ahead closed-loop interactions between EV and OV in real racing. GPR's prediction surpasses the physics-based CAV method, yet struggles to adapt to different OV driving policies, leading to a higher risk of collisions due to misapprehension of the driving policy, e.g., blocking as a non-blocking policy and vice versa. Similarly, DKL and DNN falls short of capturing the interactions between EV and OV, resulting in a higher collision rate compared to KM-DKL. In contrast, our KM-DKL, with its prediction ability against different OV driving policies, facilitates agile and safe overtaking maneuvers without major collisions. Furthermore, computational complexity is evaluated using the average computation time in milliseconds. Although our method requires more computation time compared to other baselines, it meets the 10Hz real-time requirement, making it feasible for high-speed autonomous racing.


\subsection{Ablation study for unseen driving policy}
To further investigate the virtue of our method, we examine the uncertainty calibration performance of the proposed KM-DKL model compared to the nominal DKL, the ablated version of our model. This involves evaluating the prediction model against out-of-training-distribution scenarios, specifically when the encountered OV's driving pattern is not included in the training dataset, i.e., unseen driving policies. To create the unseen driving policy, we introduce a cooperative driving policy designed to yield the pathway to the EV. We realize this policy by modifying the aggressive driving policy's cost with the reversed sign of the blocking weight. This encourages the OV to move away from the EV's anticipated path during the race.

\begin{figure}[ht]
\centering\
\includegraphics[width=0.34\textwidth]{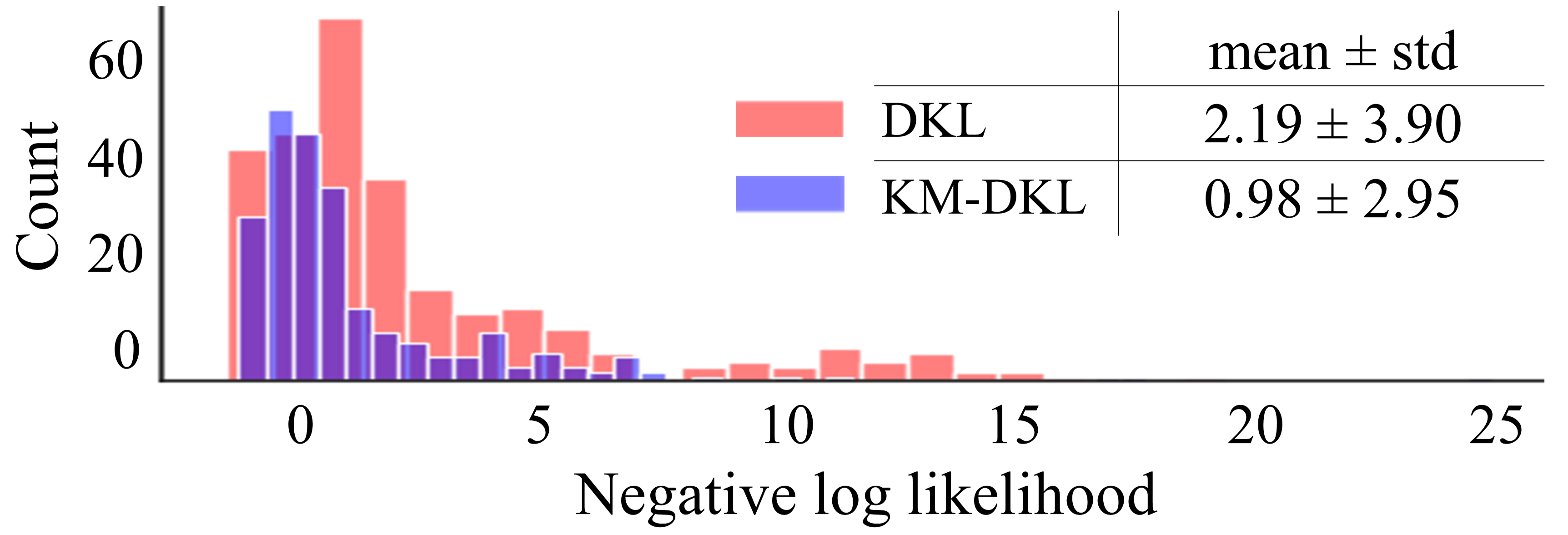}
\caption{Histograms of the negative log-likelihood for prediction results in races against OVs exhibiting unseen driving policies.\vspace{-6mm}}\label{fig:ood}
\end{figure}

We carry out the experiment in a similar fashion to the aforementioned racing setup but with the unseen OV driving policy. The prediction models, DKL and KM-DKL, are evaluated in terms of the Negative Log-Likelihood (NLL) concerning the predicted OV pose at the last prediction step. NLL quantifies the likelihood of the predicted state given the observed data, with lower values indicating more accurate uncertainty calibration to unseen data \cite{marco2023out}. Fig. \ref{fig:ood} indicates that KM-DKL achieves lower NLL values than the DKL predictor. This demonstrates our method's improved uncertainty calibration, highlighting the advantages of incorporating kernel metrics learning into DKL.

\section{Conclusions}
This study has developed a learning-based OV trajectory prediction model and downstream EV planning framework for autonomous racing against OVs with diverse driving policies. The key idea is to introduce novel heterogeneous kernel metrics and embed them into the learning pipeline of DKL. The prediction model is then learned in an unsupervised manner based on the EV-OV interaction dataset, adeptly differentiating the diverse driving policies. Compared to existing work, the proposed method enhances prediction accuracy and improves uncertainty calibration, which is instrumental for the safe and agile trajectory planning of EV. This has been validated on a 1/10th scale racecar platform, where the experimental results have shown better-calibrated prediction errors, lower collision rates, and higher overtaking success rates. Future work will focus on: i) online learning and adaptation of the algorithm to streaming data from EV-OV interactions; and ii) scaling the method to full-scale vehicles, addressing perception errors and multi-vehicle interactions in real-world racing.

\vspace{-1mm}



\addtolength{\textheight}{-0cm}   








\bibliographystyle{unsrt}
\bibliography{ref.bib}{}

@inproceedings{alahi2016social,
  title={Social lstm: Human trajectory prediction in crowded spaces},
  author={Alahi, Alexandre and Goel, Kratarth and Ramanathan, Vignesh and Robicquet, Alexandre and Fei-Fei, Li and Savarese, Silvio},
  booktitle={Proceedings of the IEEE conference on computer vision and pattern recognition},
  pages={961--971},
  year={2016}
}

@article{reiter2023frenet,
  title={Frenet-cartesian model representations for automotive obstacle avoidance within nonlinear mpc},
  author={Reiter, Rudolf and Nurkanovi{\'c}, Armin and Frey, Jonathan and Diehl, Moritz},
  journal={European Journal of Control},
  volume={74},
  pages={100847},
  year={2023},
  publisher={Elsevier}
}

@inproceedings{varadarajan2022multipath++,
  title={Multipath++: Efficient information fusion and trajectory aggregation for behavior prediction},
  author={Varadarajan, Balakrishnan and Hefny, Ahmed and Srivastava, Avikalp and Refaat, Khaled S and Nayakanti, Nigamaa and Cornman, Andre and Chen, Kan and Douillard, Bertrand and Lam, Chi Pang and Anguelov, Dragomir and others},
  booktitle={2022 International Conference on Robotics and Automation (ICRA)},
  pages={7814--7821},
  year={2022},
  organization={IEEE}
}

@inproceedings{marco2023out,
  title={Out of Distribution Detection via Domain-Informed Gaussian Process State Space Models},
  author={Marco, Alonso and Morley, Elias and Tomlin, Claire J},
  booktitle={2023 62nd IEEE Conference on Decision and Control (CDC)},
  pages={5487--5493},
  year={2023},
  organization={IEEE}
}

@article{gardner2018gpytorch,
  title={Gpytorch: Blackbox matrix-matrix gaussian process inference with gpu acceleration},
  author={Gardner, Jacob and Pleiss, Geoff and Weinberger, Kilian Q and Bindel, David and Wilson, Andrew G},
  journal={Advances in neural information processing systems},
  volume={31},
  year={2018}
}

@software{gtsam,
  author       = {Frank Dellaert and GTSAM Contributors},
  title        = {borglab/gtsam},
  month        = May,
  year         = 2022,
  publisher    = {Georgia Tech Borg Lab},
  version      = {4.2a8},
  doi          = {10.5281/zenodo.5794541},
  url          = {https://github.com/borglab/gtsam)}}

@inproceedings{hess2016real,
  title={Real-time loop closure in 2D LIDAR SLAM},
  author={Hess, Wolfgang and Kohler, Damon and Rapp, Holger and Andor, Daniel},
  booktitle={2016 IEEE international conference on robotics and automation (ICRA)},
  pages={1271--1278},
  year={2016},
  organization={IEEE}
}

@inproceedings{ober2021promises,
  title={The promises and pitfalls of deep kernel learning},
  author={Ober, Sebastian W and Rasmussen, Carl E and van der Wilk, Mark},
  booktitle={Uncertainty in Artificial Intelligence},
  pages={1206--1216},
  year={2021},
  organization={PMLR}
}

@inproceedings{wilson2016deep,
  title={Deep kernel learning},
  author={Wilson, Andrew Gordon and Hu, Zhiting and Salakhutdinov, Ruslan and Xing, Eric P},
  booktitle={Artificial intelligence and statistics},
  pages={370--378},
  year={2016},
  organization={PMLR}
}

@article{zheng2023simts,
  title={SimTS: Rethinking Contrastive Representation Learning for Time Series Forecasting},
  author={Zheng, Xiaochen and Chen, Xingyu and Sch{\"u}rch, Manuel and Mollaysa, Amina and Allam, Ahmed and Krauthammer, Michael},
  journal={arXiv preprint arXiv:2303.18205},
  year={2023}
}

@inproceedings{liang2020simaug,
  title={Simaug: Learning robust representations from simulation for trajectory prediction},
  author={Liang, Junwei and Jiang, Lu and Hauptmann, Alexander},
  booktitle={Computer Vision--ECCV 2020: 16th European Conference, Glasgow, UK, August 23--28, 2020, Proceedings, Part XIII 16},
  pages={275--292},
  year={2020},
  organization={Springer}
}

@article{gulzar2021survey,
  title={A survey on motion prediction of pedestrians and vehicles for autonomous driving},
  author={Gulzar, Mahir and Muhammad, Yar and Muhammad, Naveed},
  journal={IEEE Access},
  volume={9},
  pages={137957--137969},
  year={2021},
  publisher={IEEE}
}

@article{betz2022autonomous,
  title={Autonomous vehicles on the edge: A survey on autonomous vehicle racing},
  author={Betz, Johannes and Zheng, Hongrui and Liniger, Alexander and Rosolia, Ugo and Karle, Phillip and Behl, Madhur and Krovi, Venkat and Mangharam, Rahul},
  journal={IEEE Open Journal of Intelligent Transportation Systems},
  volume={3},
  pages={458--488},
  year={2022},
  publisher={IEEE}
}

@article{liniger2015optimization,
  title={Optimization-based autonomous racing of 1: 43 scale RC cars},
  author={Liniger, Alexander and Domahidi, Alexander and Morari, Manfred},
  journal={Optimal Control Applications and Methods},
  volume={36},
  number={5},
  pages={628--647},
  year={2015},
  publisher={Wiley Online Library}
}

@article{yoon2021interaction,
  title={Interaction-aware probabilistic trajectory prediction of cut-in vehicles using Gaussian process for proactive control of autonomous vehicles},
  author={Yoon, Youngmin and Kim, Changhee and Lee, Jongmin and Yi, Kyongsu},
  journal={IEEE Access},
  volume={9},
  pages={63440--63455},
  year={2021},
  publisher={IEEE}
}

@inproceedings{zhu2023gaussian,
  title={A Gaussian Process Model for Opponent Prediction in Autonomous Racing},
  author={Zhu, Edward L and Busch, Finn Lukas and Johnson, Jake and Borrelli, Francesco},
  booktitle={2023 IEEE/RSJ International Conference on Intelligent Robots and Systems (IROS)},
  pages={8186--8191},
  year={2023},
  organization={IEEE}
}

@article{FORCESNLP,
Author = "A. Zanelli and A. Domahidi and J. Jerez and M. Morari",
Title = "FORCES NLP: an efficient implementation of interior-point...
methods for multistage nonlinear nonconvex programs",
Journal = "International Journal of Control",
Year = "2017",
Pages = "1-17"
}

@article{wang2021game,
  title={Game-theoretic planning for self-driving cars in multivehicle competitive scenarios},
  author={Wang, Mingyu and Wang, Zijian and Talbot, John and Gerdes, J Christian and Schwager, Mac},
  journal={IEEE Transactions on Robotics},
  volume={37},
  number={4},
  pages={1313--1325},
  year={2021},
  publisher={IEEE}
}

@article{chen2022interactive,
  title={Interactive multi-modal motion planning with branch model predictive control},
  author={Chen, Yuxiao and Rosolia, Ugo and Ubellacker, Wyatt and Csomay-Shanklin, Noel and Ames, Aaron D},
  journal={IEEE Robotics and Automation Letters},
  volume={7},
  number={2},
  pages={5365--5372},
  year={2022},
  publisher={IEEE}
}

@article{feng2024unitraj,
  title={UniTraj: A Unified Framework for Scalable Vehicle Trajectory Prediction},
  author={Feng, Lan and Bahari, Mohammadhossein and Amor, Kaouther Messaoud Ben and Zablocki, {\'E}loi and Cord, Matthieu and Alahi, Alexandre},
  journal={arXiv preprint arXiv:2403.15098},
  year={2024}
}

@inproceedings{gilles2022uncertainty,
  title={Uncertainty estimation for Cross-dataset performance in Trajectory prediction},
  author={Gilles, Thomas and Sabatini, Stefano and Tsishkou, Dzmitry and Stanciulescu, Bogdan and Moutarde, Fabien},
  booktitle={ICRA 2022 Fresh Perspectives on the Future of Autonomous Driving Workshop},
  year={2022}
}


\end{document}